\documentclass[runningheads]{llncs}
\usepackage[T1]{fontenc}
\usepackage{xcolor}
\PassOptionsToPackage{table}{xcolor}

\usepackage{colortbl} 
\usepackage{graphicx}
\usepackage{multirow}
\usepackage{booktabs}
\usepackage{tabularx}
\usepackage{amsmath}
\usepackage{amsfonts}
\usepackage{amssymb}
\usepackage{pifont}
\usepackage{verbatim}
\usepackage{makecell}
\usepackage{hyperref}

\definecolor{mygray}{gray}{.9}

\usepackage{tikz}  

\newcommand{\imp}[1]{$_{{\textbf{\textcolor{Better}{#1}}}}$}

\newcommand{\worse}[1]{$_{{\textbf{\textcolor{Worse}{#1}}}}$}

\definecolor{Better}{rgb}{0.18, 0.407, 0.266}

\definecolor{Worse}{rgb}{0.55, 0.0, 0.0}

\newcommand{\xx}{\boldsymbol{x}}
\newcommand{\zz}{\boldsymbol{z}}
\newcommand{\aaa}{\boldsymbol{a}}
\newcommand{\ee}{\boldsymbol{\epsilon}}
\newcommand{\xmark}{\ding{55}}%

\newcommand{\mypar}[1]{\vspace{4pt}\noindent\textbf{#1}}

\newcommand{\Ours}{MAD-AD}

\begin{document}
\title{MAD-AD: Masked Diffusion for Unsupervised Brain Anomaly Detection}
%
%
%

\author{Farzad Beizaee \inst{1,2} \thanks{Corresponding author: \email{farzad.beizaee.1@ens.etsmtl.ca}}
\and
Gregory Lodygensky \inst{2,3}
\and
Christian Desrosiers \inst{1}
\and
\\ Jose Dolz \inst{1}
}
\authorrunning{F. Beizaee et al.}
%
\institute{
    ÉTS Montreal \and CHU Sainte-Justine Hospital, Montreal \and University of Montreal
}

\maketitle              

\begin{abstract}
Unsupervised anomaly detection in brain images is crucial for identifying injuries and pathologies without access to labels. However, the accurate localization of anomalies in medical images remains challenging due to the inherent complexity and variability of brain structures and the scarcity of annotated abnormal data. To address this challenge, we propose a novel approach that incorporates masking within diffusion models, leveraging their generative capabilities to learn robust representations of normal brain anatomy. During training, our model processes only normal brain MRI scans and performs a forward diffusion process in the latent space that adds noise to the features of randomly-selected patches. Following a dual objective, the model learns to identify which patches are noisy and recover their original features. This strategy ensures that the model captures intricate patterns of normal brain structures while isolating potential anomalies as noise in the latent space. At inference, the model identifies noisy patches corresponding to anomalies and generates a normal counterpart for these patches by applying a reverse diffusion process. Our method surpasses existing unsupervised anomaly detection techniques, demonstrating superior performance in generating accurate normal counterparts and localizing anomalies.
The code is available at \href{https://github.com/farzad-bz/MAD-AD}{hhttps://github.com/farzad-bz/MAD-AD}

\end{abstract}

\begin{figure}[h!]
\centering
\includegraphics[width=\linewidth]{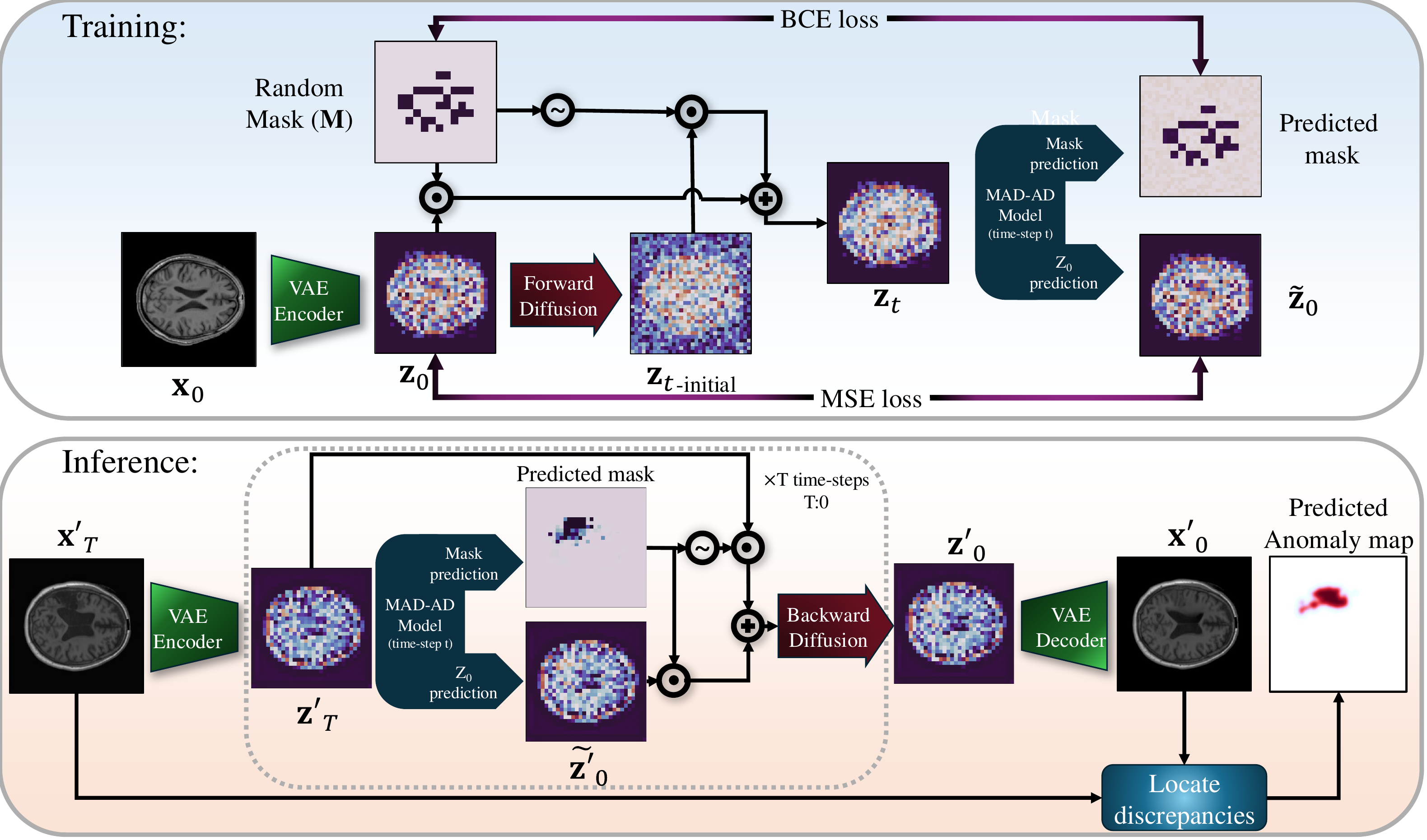}
     \caption{\textbf{Overview of the proposed method.} During \textit{training}, normal samples are encoded into latent space. A binary mask and a time-step t are applied, and non-masked regions undergo forward diffusion to produce $\zz_t$. The model is then trained to predict $\zz_0$ and the incorporated mask for forward diffusion. At \textit{inference}, the model undergoes a selective reverse process using the predicted mask at each step.}
     
\label{fig:method}
\end{figure}

\section{Introduction}

The accurate detection and localization of brain anomalies in medical images, particularly in Magnetic Resonance Imaging (MRI) data, is paramount to diagnosing and understanding neurological injuries and pathologies. 
However, the complexity of brain structures and the scarcity of labeled abnormal data present significant challenges in developing robust and generalizable solutions. 
Traditionally, brain anomaly detection has been framed as a supervised learning task, which aims at identifying well-defined pathologies such as brain tumor \cite{havaei2017brain,kamnitsas2018ensembles,sinha2020multi}, atrophy \cite{pagnozzi2019quantifying} or white matter hyper-intensities \cite{kervadec2021boundary,kuijf2019standardized}, among many others. Nevertheless, casting anomaly detection as a supervised problem introduces an inherent bias towards the targeted lesions, limiting the scope of detectable pathologies. Moreover, collecting large amounts of annotated samples encompassing the entire spectrum of potential brain abnormalities is expensive and impractical for novel structures or rare abnormal patterns.

Unsupervised anomaly detection (UAD), which involves modeling the distribution of normal data and identifying deviations as anomalies, has gained attention as a promising alternative \cite{behrendt2022unsupervised,bercea2024generalizing,silva2022constrained,zimmerer2019unsupervised}. Conventional unsupervised methods, such as autoencoders \cite{Rumelhart1986} and generative adversarial networks (GANs) \cite{goodfellow2020generative}, attempt to reconstruct normal anatomical structures and flag areas with high reconstruction errors as anomalies. Despite their potential, these approaches suffer from notable limitations. Autoencoders often fail to capture the fine-grained details of normal anatomy, whereas GANs are prone to mode collapse and instability during training. Moreover, these models frequently reconstruct anomalies as part of normal structures, reducing their reliability in clinical applications.

Recent advances in diffusion models \cite{ho2020denoising,kingma2021variational,sohl2015deep,songdenoising} have opened new avenues in generative modeling. Such models \cite{sohl2015deep} leverage a stochastic process to gradually corrupt data and learn to reverse this process, enabling them to model complex data distributions with remarkable precision. Their success in generating high-quality images and their ability to capture intricate patterns in the data have prompted researchers to explore their use for anomaly detection \cite{graham2023denoising,lu2023removing,yan2023feature,liang2023modality,naval2024ensembled}. 
While these methods have improved the accuracy of anomaly detection, 
their application to brain images introduces several challenges. Firstly, the forward diffusion process can cause a loss of distinctive features across brain regions, especially when the number of steps is large. This loss may compromise the model's ability to differentiate between normal and anomalous brain regions. Also, reducing the number of forward diffusion steps introduces the risk of an \textit{``identity shortcut''} problem. In this problem, the model can easily recover the fine details of the input image, resulting in anomalous regions being preserved in the reconstruction. This is a significant concern in brain anomaly detection, where subtle but critical deviations such as tumors or lesions may be overlooked due to this shortcut behavior. Another issue arises from the indiscriminate application of forward and reverse diffusion across the entire brain image. This approach can hinder the model's ability to effectively reconstruct normal brain patterns.

To address these limitations, we propose \Ours{}, a Masked Diffusion for brain anomaly detection with the following key contributions. First, we leverage latent diffusion models to treat anomalies as partial noise in the latent space, enabling their effective restoration through the denoising process. Our method also removes the reliance on forward diffusion steps during inference, thereby preventing the loss of critical visual details and enabling highly accurate reconstructions of the underlying normal appearance. This is accomplished by masking the forward diffusion process and training the model to reverse it effectively. Furthermore, we incorporate a mask-prediction module into the diffusion framework, allowing the prediction of the incorporated mask in the diffusion process. This approach ensures the selective correction of anomalous regions while preserving normal regions intact, ultimately delivering more precise and reliable anomaly detection results. The overview of our method is depicted in Fig. \ref{fig:method}.

\section{Related works} 

Recent approaches for unsupervised anomaly detection (UAD) in brain MRI can mainly be divided in three categories: methods based on different variants of autoencoders (AEs), those using generative adversarial networks (GANs) and the ones based on diffusion models.

\mypar{AE-based methods.~} Approaches in this category train an autoencoder on normal data to accurately reconstruct input images. At inference, the reconstruction error measured at each pixel is used to localize anomalies. Different networks have been explored for the reconstruction, including standard autoencoders (AE) \cite{atlason2019unsupervised,baur2021autoencoders}, variational autoencoders (VAEs) \cite{baur2021autoencoders,silva2022constrained,zimmerer2019unsupervised} and denoising autoencoders (DAEs) \cite{kascenas2022denoising}. A common issue with these methods is their propensity to overfit the training data, leading to a poor generalization on unseen data. Furthermore, they are prone to blurry reconstructions, struggling to accurately distinguish subtle anomalies from normal variations, especially when relying solely on reconstruction error as a measure of abnormality.

\mypar{GAN-based methods.~} These approaches employ an adversarial learning strategy where a generator and a discriminator are jointly trained on healthy subject images to learn a latent representation of normal variability. AnoGAN~\cite{schlegl2017unsupervised} measures anomaly scores based on a combination of reconstruction error and distance in the latent space. f-AnoGAN~\cite{schlegl2019fast} improves upon this work by incorporating an additional feature-level reconstruction strategy, yielding a more precise localization of anomalies. The work in~\cite{baur2020steganomaly} uses a style transfer method based on CycleGANs to map real MR images of healthy brains to synthetic ones, and vice versa. Anomalies are then detected by comparing input images to their reconstruction. While the ability of GANs to generate high-quality images can translate in a more detailed delineation of anomalies, they are also prone to training instability and are often sensitive to hyperparameter choices.

\mypar{Diffusion-based method.~} 
Diffusion models have gained significant attention in computer vision for their ability to generate high-fidelity images \cite{croitoru2023diffusion}. Recently, these models have also shown promise in various medical image analysis tasks including UAD~\cite{behrendt2023patched,iqbal2023unsupervised,behrendt2024leveraging,bercea2024diffusion,liang2024itermask2,naval_marimont2024disyre,naval2024ensembled,wyatt2022anoddpm}. A prominent diffusion-based method for UAD in medical images,  
AnoDDPM~\cite{wyatt2022anoddpm} utilizes a partial diffusion strategy, adding noise to an image up to a specific timestep and then recovering the original image with a reverse diffusion process. This method has shown success in detecting anomalies in brain MRI and other domains.   PDDPM~\cite{behrendt2023patched} instead applies the diffusion process in a patch-wise manner, aiming to improve the understanding of local image context and achieve better anatomical coherence in the reconstruction. This method divides the image into overlapping patches and reconstructs each patch while considering its unperturbed surroundings. CDDPM~\cite{behrendt2024leveraging} generates multiple reconstructions via the reverse diffusion process and pinpoints anomalies by examining the distribution of these reconstructions with the Mahalanobis distance, subsequently labeling outliers as anomalies. MDDPM\cite{iqbal2023unsupervised} incorporates masking-based regularization, applied on both image patches and in the frequency domain, to enhance unsupervised anomaly detection. AutoDDPM \cite{bercea2023mask} incorporates automatic masking, stitching, and resampling techniques within the DDPM framework to enhance its robustness and accuracy in anomaly detection. This approach also addresses the challenge of selecting an appropriate noise level for detecting lesions of various sizes. However, the diffusion-based UAD models mentioned above rely heavily on a forward diffusion process that inherently results in information loss. Consequently, these methods often fail to accurately reconstruct the original healthy brain structures, leading to false-positive detections where normal regions are incorrectly identified as anomalous. This issue is particularly prominent in brain anomaly detection tasks, as brain structures, especially cortical regions, vary uniquely across individuals, thereby increasing the difficulty of accurately recovering normal anatomical variations.

A recently proposed method, DISYRE \cite{naval_marimont2024disyre,naval2024ensembled}, uses a diffusion-like pipeline to train a model to restore images that have been corrupted with synthetic anomalies. Anomalies in a new image are detected based on the model's ability to restore the image to a healthy state. A key limitation of this method is that the synthetic anomalies may not encompass all types of real-world anomalies, limiting its generalization ability. THOR \cite{bercea2024diffusion} integrates implicit guidance into the DDPM's denoising process using intermediate masks to preserve the integrity of healthy tissue details. It aims to ensure a faithful reconstruction of the original image in areas unaffected by pathology, minimizing false positives. However, since these intermediate masks are determined based on the perceptual differences between input images and their reconstruction at each step, the model may struggle to detect subtle or small anomalies, as they might be masked out due to their minimal differences with the input image. Additionally, reconstruction errors may occur due to the loss of details during the forward process, with normal regions not getting masked due to their high perceptual differences. Inspired by diffusion-based models, IterMask$^2$ \cite{liang2024itermask2} incorporates an iterative spatial mask refinement process and frequency masking to enhance UAD performance. This strategy minimizes information loss in normal areas by iteratively shrinking a spatial mask, starting from the whole brain towards the anomaly. Although the model performs well in detecting hypo- or hyper-intense areas, it can fail to localize structural anomalies such as atrophy or enlarged ventricles as their reconstruction is conditioned on structural information from high-frequency image components which can be recovered by the model.

\section{Method}
\subsection{Modeling the normal feature space} 
We resort to diffusion models for learning the space of normal data and reconstructing the normal counterpart of anomalous regions. Denoising Diffusion Probabilistic Models (DDPMs) \cite{ho2020denoising} learn a data distribution by gradually adding noise to the data (i.e., forward process) and then training a model to reverse this process. While DDPMs are highly effective at generating high-quality images, there are certain limitations when using them directly for detecting anomalous regions. Firstly, the number of steps in the forward diffusion process can have a considerable impact on the performance. If this number is too large, semantic information of the brain structure can be lost, resulting in an uncorrelated brain reconstruction and the incorrect detection of normal regions as abnormal. On the other hand, if not enough steps are used, the model can too easily recover the fine details in the image. As a result, abnormal regions will incorrectly be detected as normal. Moreover, as normal patches are also affected by noise, they cannot be fully exploited to reconstruct abnormal regions. 

To overcome the aforementioned limitations, we propose to incorporate a random masking strategy in the diffusion model and modify the reverse process so that the diffusion model can selectively alter anomalous parts of an image, while keeping the normal regions untouched. Following \cite{rombach2022high,peebles2023scalable}, we employ a diffusion model operating in the \emph{latent} space. This has two important advantages. First, whereas adding Gaussian noise directly on the image yields corruptions that have no meaningful structure, injecting this noise on latent features and then reconstructing these noisy features results in more complex corruptions that better represent real anomalies in brain MRI.  Moreover, this also mitigates the ``identity shortcut'' problem, enhances computational efficiency, and improves stability, particularly with limited training data. 

Let $\mathcal{X}=\{\xx^{(i)}\}^N_{i=1}$ be the training set consisting exclusively of normal 
images $\xx^{(i)} \in \mathbb{R}^{H \times W \times C}$, 
where $H$, $W$, and $C$ correspond to the image height, width, and number of channels, respectively. We employ a pre-trained variational autoencoder~\cite{rombach2022high}, which is adapted and fine-tuned for medical images. This model can encode high-dimensional image data into a compact latent representation and reconstruct this data from the latent space while preserving essential structural and semantic information. Denoting the encoder network as $\mathrm{V}_{E,\phi}$, an input image $\xx^{(i)}$ is mapped to its latent space representation $\zz^{(i)}=\mathrm{V}_{E,\phi}(\xx^{(i)})$, where $\zz^{(i)} \in \mathbb{R}^{H' \times W' \times C'}$. 

\mypar{Random masking.~} To incorporate random masking into the forward diffusion process, given the latent features of an input normal sample $\zz_0 \sim p\left(\zz_0\right)$, we spatially divide $\zz_0$ into non-overlapping patches defined by a random mask $M \in [0, 1]^{H \times W}$. The forward Markov diffusion process to generate samples $\zz_t$ gradually applies noise to the non-masked patches of sample $\zz_0$ for $t$ time steps, where $t \in [1, T]$. Following \cite{ho2020denoising}, the forward noising process in the latent space with masking can be characterized as: 
\begin{equation}
\zz_t = \left(\sqrt{1 - \beta_t} \zz_{t-1} + \sqrt{\beta_t} \boldsymbol{\epsilon}\right) \odot M + \zz_0 \odot (1 - M),
\end{equation}
where $\zz_t$ is the partially diffused image at step $t$, $\epsilon \sim \mathcal{N}(0, I)$ is the sampled Gaussian noise and $\beta_t$ is the noise schedule at step $t$, which controls the amount of noise added at each step. Using the reparameterization trick, $\zz_t$ can be obtained implicitly using the following equation:
\begin{equation}
\label{eq:masking}
\zz_t = \left(\sqrt{\bar{\alpha}_t} \zz_0 + \sqrt{1 - \bar{\alpha}_t} \boldsymbol{\epsilon}\right) \odot M + \zz_0 \odot (1 - M),
\end{equation}
with $\alpha_t=1-\beta_t$ and  $\bar{\alpha}_t=\prod_{i=1}^T \alpha_i$. The reverse process aims to recover the original data $\zz_0$ by gradually removing the noise. This process is modeled as a learned distribution that reverses the forward noising steps. Given the masked sample $\zz_t$ at step $t$ and mask $M$ at spatial location $k$, the reverse process can be modeled as follows:
\begin{equation}
p(\zz_{t-1}^k|\zz_{t}^k)\,=\,
\left\{ 
  \begin{array}{l}
    \mathcal{N}(\zz_{t-1}^k;\mu_{\theta}(\zz_t^k,t),\ \beta_{t} \mathbf{I}), \ \ \textsl{if $M^k=1$}  \\[5pt]
    \zz_{t}^k, \ \ \hspace{3cm} \ \textsl{otherwise;} 
  \end{array}
\right.
\label{eq:prior}
\end{equation}
In this equation, 
$\mu_{\theta}(\zz_t,t)$ is a trainable function, which can be reparameterized as a predicted noise $\epsilon$ or a predicted clean image $\zz_0$. Due to the incorporated random masking strategy, we prefer the latter one for simplicity. Therefore, $\mu_{\theta}(\zz_t,t)$ can be formally expressed as:

\begin{equation}
\label{mu}
\mu_{\theta}(\zz_t,t)\,=\,\frac{\sqrt{\bar{\alpha}_{t-1}} \beta_t}{1-\bar{\alpha}_t} f_{\theta,\zz_0}(\zz_t, t) \, + \,\frac{\sqrt{\alpha_t}\left(1-\bar{\alpha}_{t-1}\right)}{1-\bar{\alpha}_t} \zz_t,
\end{equation}
where $f_{\theta,\zz_0}(\zz_t, t)$ is a function that predicts $\tilde{\zz}_0$ at time step t, given $\zz_t$.  

\mypar{Mask prediction.~} By parameterizing $f_{\theta}$ as a neural network, the model can be trained using a simple mean-square error loss between $\zz_0$ and the predicted clean image. Moreover, in Eq. (\ref{eq:prior}), we  assumed that the mask $M$ is available in the reverse process. However, this assumption is unrealistic since the mask used in diffusing the image, which contains the location of anomalous regions, is not accessible at inference. Therefore, we 
include an additional head $f_{\theta,M}$ to the diffusion model that predicts the mask used in the forward diffusion. This can be achieved by applying a binary cross-entropy ($\mathcal{L}_{BCE}$) loss between the predicted mask from this head and a randomly sampled mask used during partial diffusion in training. The final training objective of our model is defined by:
\begin{equation}
\min_\theta \ \ \mathbb{E}_{\zz_0 \sim q\left(\zz_0\right), \ee, t, M}\left[\left\|\zz_0-f_{\theta,\zz_0}\left(\zz_t, t\right)\right\|_2^2 \, + \, \lambda\mathcal{L}_{BCE}\big(M,\,f_{\theta,M}(\zz_t, t)\big)\right],
\end{equation}
where $\lambda$ is a hyper-parameter that balances the contributions of the two terms. 

\subsection{Recovering normal images} 

During inference, the goal is to recover a normal version of an abnormal brain image, where anomalous regions are replaced with their normal counterpart while normal areas remain unchanged. As previously discussed, a pre-trained VAE, $V(\cdot)$, is employed to project the image into a latent space where the data follows a normal distribution. In this space, abnormal brain regions can be interpreted as normal noise, as they fall outside the learned normal distribution of the model. These abnormal areas can also be considered as non-masked regions through the forward diffusion process using a mask that points out anomalous regions. Consequently, the proposed method incorporates all the necessary components to first predict the location of anomalies using the mask prediction head and then progressively denoise these regions to reconstruct their normal counterpart. Finally, by comparing the input image with its corrected version, anomalies can be accurately localized. The following section provides a detailed explanation of the sampling process in the \Ours{} model during inference.

Let $\mathcal{X}'=\{\xx'^{(i)}\}^{N'}_{i=1}$ denote the test set at inference time, which consists of samples with potential anomalies. We first map these images into the latent space using $\mathrm{V}_{E, \phi}$. As explained before, we treat the latent space of an anomalous image as step $T$ of the masked forward diffusion process applied on its normal counterpart, i.e., $\zz'_{T}=\mathrm{V}_{E,\phi}(\xx'_{T})$. By predicting the mask that corresponds to the anomaly location and the reconstructed $\Tilde{\zz}'_0$ at each time-step $t$, using Eq.~(\ref{eq:prior}), we can progressively correct the anomaly regions and obtain the normal counterpart ($\zz'_{T}\rightarrow\zz'_{0}$) while preserving fine details of the normal regions. 

\begin{figure}[t!]
    \centering
    \includegraphics[width=0.8\linewidth]{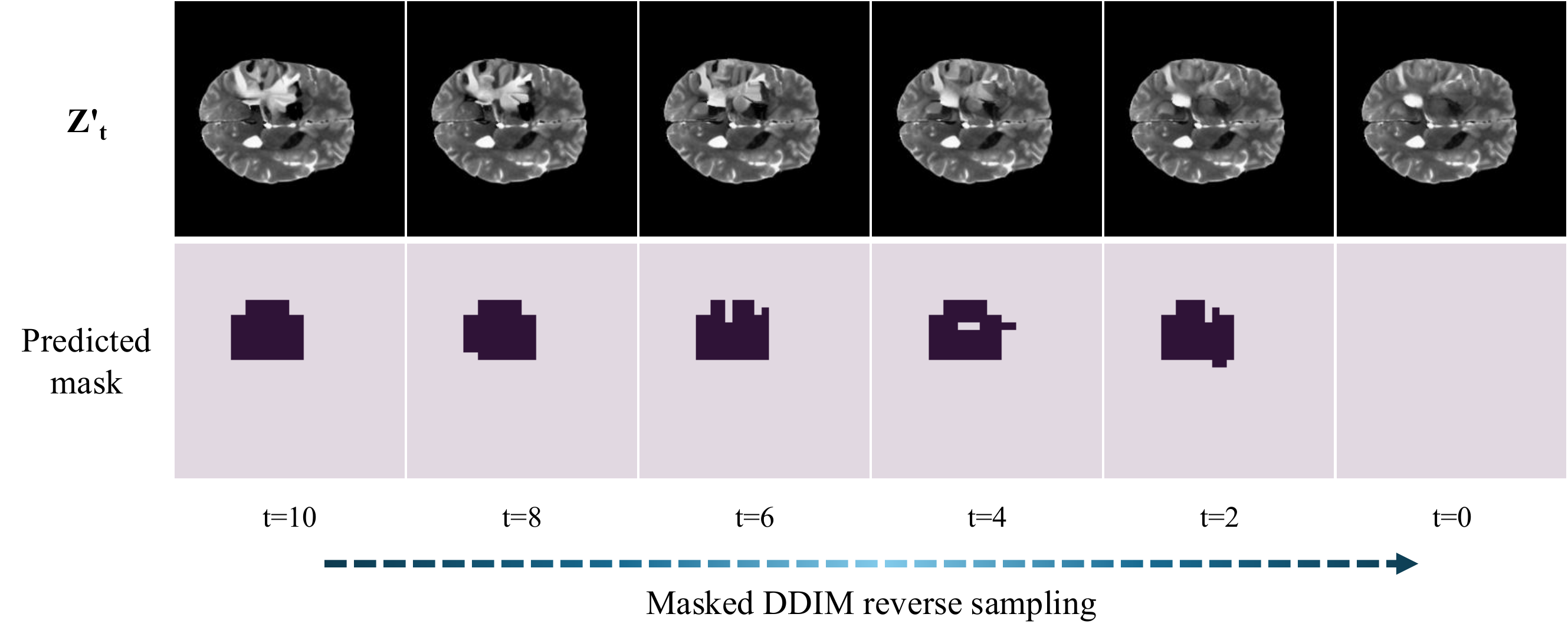}
    \caption{\textbf{Visual example of the reverse process.} Both the predicted mask and the decoded latent representation of the intermediate reverse step ($\zz'_t$) at multiple time steps are depicted to highlight the masked reverse sampling in \Ours.}
    \label{fig:reverse}
\end{figure}

Nevertheless, one drawback of sampling with DDPM is that it requires many reverse sampling steps to obtain the normal version. Therefore, we instead opted for DDIM \cite{songdenoising} which, by reducing the stochasticity of DDPM, makes the reverse process more deterministic and requires fewer sampling steps. Consequently, we modify the reverse process of DDIM for the \Ours{} model as:
\begin{equation}
\label{sampling-DDIM}
\begin{aligned}
\zz'_{t-1} =\ & \underbrace{B\Big(f_{\theta,M}(\zz'_t)\Big)}_{\textrm{``predicted mask''}} \Big(\sqrt{\bar{\alpha}_{t-1}}\underbrace{f_{\theta, z_0}(z'_t)}_{\textrm{``predicted $\tilde{\zz}'_0$''}}+\underbrace{\sqrt{1-\bar{\alpha}_{t-1}} \tilde{\ee}_{t}(\zz'_t)}_{\textrm{``direction pointing to $\zz'_t$''}} +\ \sigma_{t}\ee'_{t} \Big) \\ 
& + \Big(1 - B\big(f_{\theta,M}(\zz'_t)\big)\Big) \cdot \zz'_{t}
\end{aligned}
\end{equation}
where $B(.)$ is a binarization function, $\ee'_{t}$ is random normal noise, $\sigma_{t}$ is a hyper-parameter that controls the stochasticity of reverse process, and $\tilde{\ee}$ is the predicted noise calculated based on the predicted $\tilde{\zz}_{0}$ and $\zz_{t}$ as follows:
\begin{equation}
\tilde{\ee_{t}} \, = \, \frac{\zz'_t \cdot f_{\theta, \zz'_0}(\zz'_t)}{\sqrt{1-\bar{\alpha_{t}}}} 
\end{equation}
As mentioned above, $\sigma_t$ controls the noise level and stochasticity of the sampling process in DDIMs. Specifically, $\sigma_t=0$ makes the model deterministic, while $\sigma_t~>~0$ introduces stochasticity. For $\sigma_t=1$, the model behaves like a DDPM, 
where the sampling process involves full stochasticity with noise added at each step.
While having a fully deterministic model can be desirable for UAD 
applications, introducing a bit of noise to the non-masked (anomalous) regions helps bring the distribution closer to normal. 
This makes it easier for the model to recover the normal variation of the input. Therefore, we propose to use an in-between value of $\sigma_t=0.5$. 
A qualitative example of the reverse process in \Ours{} is depicted in Figure \ref{fig:reverse}.

\subsection{Anomaly localization}

Equation \ref{sampling-DDIM} enables a correcting trajectory from $\zz'_{T}$ to $\zz'_0$, resulting in generating high-quality normal variation of the anomalous image in fewer steps. To accurately localize anomalies, we used the discrepancy between the input image and its reconstructed normal counterpart. More concretely, using the ``normal'' latent embedding $\tilde{\zz}'_0$, we generated a reconstructed normal sample in the image-space as $\tilde{\xx}'_{0} = \mathcal{V}_{D,\phi}(\tilde{\zz}'_{0})$, where $\mathcal{V}_{D,\phi}$ is the pre-trained VAE decoder. The predicted anomaly map is then given by:
\begin{equation}
    \aaa \, = \,  G  \ast \min(\|\tilde{\xx}'_{0} - \xx'_{0}\|, \gamma)/\gamma,
\end{equation}
where $G$ is a Gaussian kernel to smooth the predicted mask, $\ast$ is the convolution operator, and $\gamma$ is a threshold designed to prevent assigning excessive weight to patches with significant deviations.

\section{Experiments}

\subsection{Experimental setting}\label{sec:setting}

\mypar{Datasets.~} We employ three datasets to asses the performance of UAD methods. \textbf{IXI Dataset} \cite{ixi}: a publicly available resource with brain MRI scans from approximately 600 healthy subjects. \textbf{ATLAS 2.0} \cite{liew2022large} includes 655 T1-weighted MRI scans accompanied by expert-segmented lesion masks. As a pre-processing, all brain scans of both IXI and ATLAS 2.0 datasets were registered to MNI152 1mm templates and normalized to the 98th percentile. Then, mid-axial slices were extracted and padded to the resolution of 256$\times$256 pixels. \textbf{BraTS'21} \cite{bakas2017advancing}: following the experimental setup of IterMask$^2$~\cite{liang2024itermask2}, we also employ this dataset, which comprises 1251 brain scans across four modalities: T1-weighted, contrast-enhanced T1-weighted (T1CE), T2-weighted, and T2 Fluid Attenuated Inversion Recovery (FLAIR). 
For each scan, 20 middle axial slices of the skull-stripped brain are extracted, which are padded to the resolution of 256$\times$256 pixels.

\mypar{Training/Testing protocol.~} We found that the training and testing protocols considerably differ in the UAD literature. For a fair comparison with prior methods, we evaluated our approach in two widely-adopted settings, comparing against the methods that were originally evaluated in each of these settings. \textbf{Setting-1 (\textit{S1})} \cite{bercea2024diffusion}: training is performed on the middle slices of IXI subjects, whereas only middle slices of ATLAS 2.0 are used for testing. 
\textbf{Setting-2 (\textit{S2})} \cite{liang2024itermask2}: in this setting, only normal slices from a given modality are used for training, while the abnormal slice of that modality with the largest pathology is employed for inference. The BRATS'21 dataset is used in this case, which is split into training (80\%), validation (10\%) and testing (10\%).

\mypar{Evaluation metrics.~} To evaluate the performance of our brain anomaly detection model, we use the Maximum Dice score, which reports the highest value obtained for thresholds ranging from 0 to 1.
Following \cite{bercea2024diffusion}, we employ the global Maximum Dice score in setting \textit{S1}, which first flattens and concatenates all segmentations and predictions before calculating the maximum Dice score. For setting \textit{S2}, we instead consider the regular Maximum Dice score. 

\mypar{Implementation Details.~} To project the data into the latent space, we employed a pre-trained perceptual compression VAE model \cite{rombach2022high}. This model leverages an autoencoder trained using a combination of perceptual loss \cite{zhang2018unreasonable} and a patch-based adversarial objective, allowing for effectively reducing the spatial dimension by a factor of 8 ($256\rightarrow32$). As this model was originally trained for RGB images, we further adapted it and fine-tuned it for single-channel brain MRI data. Then, the VAE remained frozen throughout training the diffusion model (we used a UNet with attention as the diffusion model). The number of training and inference time-steps ($T$) is set to 10. To form the random mask at each iteration, the masking ratio is drawn from a uniform distribution $U[0, 0.4]$, and the patch sizes of the mask along the X and Y axes are sampled independently from the following set: $\{1,2,4,8\}$. The random mask is then multiplied by the brain mask to prevent noise in non-brain (i.e., background) patches. The model was trained for 300 epochs using a batch size of 96 and AdamW optimizer with a learning rate of $5\times 10^{-4}$.

\subsection{Results}

\setlength{\tabcolsep}{5pt}%
\begin{table}[t!]
\centering
\caption{\textbf{Performance in setting \textit{S1:}} results across different lesion sizes, where bold
highlights the best method and improvements of our approach compared to the best baseline are indicated in green.} 
\resizebox{.9\columnwidth}{!}{
\begin{tabular}{l|cccc}
\toprule
\label{table:set1}
\multirow[b]{2}{*}{Method} &  \multicolumn{4}{c}{Pathology (Global Max Dice)$\uparrow$}\\
 \cmidrule(l{5pt}r{5pt}){2-5}
 &  Average & Small &  Medium &  Large \\
 \midrule
DDPM~\cite{ho2020denoising}$_{\tiny\textsl {NeurIPS'20}}$  & 8.1 & 1.4 & 9.5 & 25.7 \\
AnoDDPM~\cite{wyatt2022anoddpm}$_{\tiny\textsl {CVPRw'22}}$ & 18.1 & 4.8 & 23.5 & 46.7 \\
AutoDDPM~\cite{bercea2023mask}$_{\tiny\textsl {ICMLw'23}}$ & 17.0 & 4.5 & 22.1 & 43.5 \\
pDDPM~\cite{behrendt2023patched}$_{\tiny\textsl {MIDL'24}}$ & 22.3 & 8.0 & 30.2 & 47.7 \\
THOR~\cite{bercea2024diffusion}$_{\tiny\textsl {MICCAI'24}}$ & 29.7 & 11.5 & 39.2 & 63.6 \\
\rowcolor{mygray} 
\textbf{\Ours} \textit{(Ours)} & \textbf{51.6}\imp{+21.9} & \textbf{15.5}\imp{+4.0} & \textbf{50.1}\imp{+10.9} & \textbf{64.1}\imp{+0.5} \\
\bottomrule
\end{tabular}
}

\end{table}

\mypar{Quantitative results.~} We empirically validated our method against a set of relevant state-of-the-art brain unsupervised anomaly detection methods in the two settings described in Section \ref{sec:setting}. Table \ref{table:set1} reports the results under the first setting, which uses middle slices of the IXI dataset for training, and middle slices of ATLAS 2.0 for evaluation. We can observe that the proposed approach substantially outperforms existing diffusion-based methods, particularly on small- and medium-sized lesions. More concretely, our approach improves the best baseline (the recent THOR method \cite{bercea2024diffusion}) by 4.0\% and 10.9\% in small and medium lesions, respectively, and by 21.9\% when using the whole dataset (referred to as \textit{``Average''}, as in \cite{bercea2024diffusion}). The performance gap further increases if we consider the second best baseline (i.e., pDDPM), where average differences are equal to nearly 30\%. Note that even though our model yields superior performance for small pathologies, it still struggles to accurately locate these type of small abnormalities, similarly to existing approaches. In \Ours{}, this low performance may be due to the use of a diffusion model on a compressed latent space, which can lead to overlooking very small pathologies.

\begin{table}[t!]
\centering
    \caption{\textbf{Performance in setting \textit{S2:}} results across different modalities, where bold
highlights the best method and performance improvements (\textit{resp.} decrease) of our approach compared to the best baseline are indicated in \textcolor{Better}{\textbf{green}} (\textit{resp.} \textcolor{Worse}{\textbf{red}}).}
\label{table:set2}
\setlength{\tabcolsep}{7pt}
\resizebox{.9\columnwidth}{!}{
\begin{tabular}{l|ccccc}
\toprule 
\multirow[b]{2}{*}{Method} &  \multicolumn{5}{c}{Modality (Max Dice)$\uparrow$}\\
\cmidrule(l{5pt}r{5pt}){2-6}
 & FLAIR & T1CE & T2-w & T1-w & Avg\\
\midrule
AE~\cite{baur2021autoencoders}$_{\tiny\textsl {MedIA'21}}$ & 33.4 & 32.3 & 30.2 & 28.5 & 31.1 \\
DDPM~\cite{ho2020denoising}$_{\tiny\textsl {Neurips'20}}$ & 60.7 & 37.9& 36.4  & 29.4 & 41.1 \\
AutoDDPM~\cite{bercea2023mask}$_{\tiny\textsl {ICMLw'23}}$ & 55.5 & 36.9 & 29.7 & 33.5 &  38.9 \\
Cycl.UNet~\cite{liang2023modality}$_{\tiny\textsl {MICCAI'23}}$ & 65.0 & 42.6 & 49.5 & 37.0 & 48.5 \\
DAE~\cite{kascenas2022denoising}\scriptsize$[0, \infty]$$_{\tiny\textsl {MIDL'22}}$ & 79.7  & 36.7  & 69.6 & 29.5 & 53.9\\
IterMask$^2$~\cite{liang2024itermask2}$_{\tiny\textsl {MICCAI'24}}$ & \textbf{80.2} & 61.7 & 71.2 & 58.5 & 67.9\\
\rowcolor{mygray}
\textbf{\Ours} \textit{(Ours)} & 76.2\worse{-4.0} & \textbf{68.5}\imp{+6.8} & \textbf{73.2}\imp{+2.0} &  \textbf{63.4}\imp{+4.9} & \textbf{70.3}\imp{+2.4}\\
\bottomrule
\end{tabular}
}
\end{table}

Under the second setting (\textit{S2}), the proposed approach yields the best scores in three out of four modalities, leading to the highest average score (Table \ref{table:set2}). While the differences with respect to the best baseline are smaller in this setting, improvements over the second best baseline are still considerably high, with an overall boost near to 16\%. Thus, quantitative results under two common settings in the UAD literature demonstrate the superior performance of our approach for this task, highlighting its potential as a powerful alternative to existing methods.

\mypar{Ablation on using different sources for the anomaly score.}
In this section, we investigate the impact of using different strategies to form the anomaly map: pixel-level discrepancies ($\xx'_{0}, \xx'_{T} $), latent-space discrepancies ($\zz'_{0}, \zz'_{T} $), and the average of the predicted mask at reverse diffusion steps ($\frac{1}{T}\sum_{t=1}^{T} f_{M,\theta}(\zz'_{t})$). These results, which are reported in Table \ref{tab:anomly-map-source}, showcase the better performance of resorting to the image-level difference, motivating our design choice.

\begin{table}[h!]
    \centering
\caption{Effect of different sources for the anomaly score in \Ours{} (BRATS'21).}
\label{tab:anomly-map-source}
\resizebox{.75\columnwidth}{!}{
\begin{tabular}{l|ccccc}
\toprule
\multirow[b]{2}{*}{Anomaly source} &  \multicolumn{5}{c}{Modality (Max Dice)$\uparrow$}\\
\cmidrule(l{5pt}r{5pt}){2-6}
 & T1-w & T1CE & T2-w & FLAIR & Avg \\
 \midrule
Average predicted mask & 60.4 & 62.3 & 65.6 & 66.1 & 63.6 \\
Latent-level diff & 63.0 & 66.2 & 69.6 & 75.5 & 68.6 \\
Image-level diff & \textbf{63.4} & \textbf{68.5} & \textbf{73.2} & \textbf{76.2} & \textbf{70.3}\\
\bottomrule
\end{tabular}
}
\end{table}

\mypar{Impact of hyper-parameters.~} Next, we evaluate the influence of key hyper-parameters on the performance of the proposed method, whose results on the BRATS dataset are depicted in Table \ref{tab:ablation-hyperparameter}. From these results, we can observe that the choices made for the hyper-parameters lead to the best results overall.

\begin{table}[h!]
    \centering
\caption{Ablation study on two key hyper-parameters of \Ours.}
\label{tab:ablation-hyperparameter}
\resizebox{.75\columnwidth}{!}{
\begin{tabular}{cc|ccccc}
\toprule
\multirow[b]{2}{*}{\makecell[c]{Hyper- \\ parameter}} & \multirow[b]{2}{*}{Value} & \multicolumn{5}{c}{Modality (Max Dice)$\uparrow$}\\
\cmidrule(l{5pt}r{5pt}){3-7}
&  & T1-w & T1CE & T2-w & FLAIR & Avg \\
 \midrule
\midrule
\multirow{3}{*}{\#DDIM steps} & 2 & 62.3 & \textbf{70.1} & 68.5 & 75.3 & 69.0 \\
 & 5 & 63.1 & 69.4 & 71.1 &  74.0 & 69.4 \\
& 10 & \textbf{63.4} & 68.5 & \textbf{73.2} & \textbf{76.2} & \textbf{70.3} \\
\midrule
\multirow{3}{*}{$\gamma$} & 0.2 & \textbf{63.4} & \textbf{68.5} & 73.2 & \textbf{76.2} & 70.3\\
& 0.4 & 63.3 & 68.3 & \textbf{73.8} & 76.0 & \textbf{70.3}\\
& \xmark & 62.0 & 67.9 & 72.6 & 74.9 & 69.3 \\
\bottomrule
\end{tabular}
}

\end{table}

\begin{figure}[t!]
    \centering
    \includegraphics[width=\linewidth]{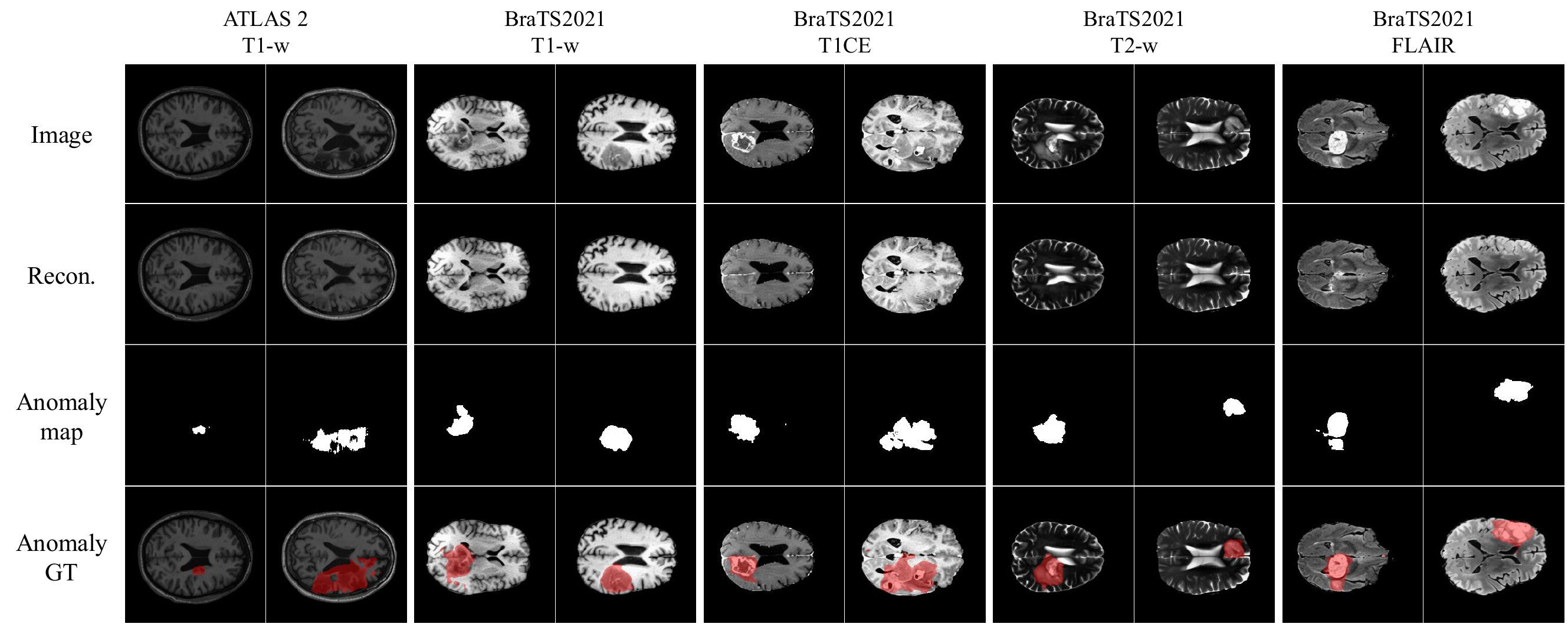}
    \caption{\textbf{Qualitative results.}  Anomaly segmentation performance obtained by our approach (i.e., ``Anomaly map") in brain MRI for different modalities and datasets.}
    \label{fig:qualitative}
\end{figure}

\mypar{Qualitative results.~} 
To further highlight the effectiveness of our unsupervised anomaly detection method, we present qualitative results obtained on the ATLAS 2.0 dataset (\textit{S1}) and across all modalities of the BraTS dataset (\textit{S2}). Figure \ref{fig:qualitative}. Figure \ref{fig:qualitative} showcases representative examples of anomalous instances, their normal counterpart reconstructions, segmentation, and anomaly map by \Ours. These qualitative results underscore the ability of our approach to accurately localize anomalous regions without relying on supervised labels.

\section{Conclusion}
This paper introduces a novel unsupervised anomaly detection method for brain MRI using a latent diffusion model with a random masking strategy. The approach leverages latent space, as brain anomalies in the latent space could be considered as noise and therefore be removed during the denoising process of diffusion models. Furthermore, by using a mask prediction module in the diffusion model, the model can selectively modify anomalous regions while preserving normal areas, enabling accurate identification of anomalous regions. Experiments on two datasets and two common brain UAD experimental settings demonstrate the superiority of our approach, validating its effectiveness in detecting and localizing brain anomalies without requiring labeled data, and showcasing its promising potential as an alternative to existing methods.

\newpage

%
%
%
\bibliographystyle{splncs04}
\bibliography{main.bib}
%





\end{document}